\documentclass{article}

\usepackage{listings}
\usepackage{graphicx}
\usepackage{multirow}
\usepackage{xcolor}

\usepackage{rotating}
\usepackage{tikz}
\usepackage{wrapfig}
\usepackage{adjustbox}
\usepackage{tabularx}
\usetikzlibrary{shapes,positioning,decorations.pathmorphing}
\usetikzlibrary{decorations.pathmorphing}
\newcommand{\thinkB}[1]{%
  \begin{tikzpicture}
    \node [rectangle, draw, decoration={bumps, path has corners, amplitude=-2.5pt, segment length=15pt}, decorate, align=center, inner sep=3mm] {#1};
  \end{tikzpicture}%

}
\usepackage{PRIMEarxiv}

\usepackage[utf8]{inputenc} 
\usepackage{hyperref}       
\usepackage{url}            
\usepackage{booktabs}       
\usepackage{amsfonts}       
\usepackage{nicefrac}       
\usepackage{microtype}      
\usepackage{lipsum}
\usepackage{fancyhdr}       
\usepackage{graphicx}       
\graphicspath{{media/}}     

\title{ Conversational Ontology Alignment with ChatGPT
}

\author{
  Sanaz Saki Norouzi, Mohammad Saeid Mahdavinejad, Pascal Hitzler \\
  Department of Computer Science \\
  Kansas State University \\
  \texttt{\{sanazsn, saeid, hitzler\}@ksu.edu} \\
}

\begin{document}
\maketitle

\large
\begin{abstract}
This study evaluates the applicability and efficiency of ChatGPT for ontology alignment using a naive approach. ChatGPT's output is compared to the results of the Ontology Alignment Evaluation Initiative 2022 campaign using conference track ontologies. This comparison is intended to provide insights into the capabilities of a conversational large language model when used in a naive way for ontology matching, and to investigate the potential advantages and disadvantages of this approach.
\end{abstract}

\keywords{Ontology alignment \and ChatGPT\and Schema matching \and Ontology matching \and Large language models\and
  Prompt engineering\and
  LLM behaviour}

\section{Introduction}

Ontology alignment (OA), also referred to as ontology matching, is a central task in semantic web technologies
that aims to find semantic correspondences between two ontologies with overlapping domains.
As using ontologies is extending to many different fields, this task’s importance is increasing, so ontology matching is required for bridging the semantic gap between various ontologies \cite{shvaiko2011ontology}.
Although OA already looks back to many years of research, the task remains challenging, often requiring expert intervention
to ensure accurate results. Expert-driven matching can be both time-consuming and subject to
human biases, so even in this case absolute precision remains elusive \cite{trojahn2022foundational,stevens2019measuring,DBLP:conf/semweb/CheathamH14}.
To tackle this challenge, a variety of ontology matching systems, incorporating natural language processing (NLP) techniques considering grammar changes and different similarity measurements, machine learning, fuzzy lexical
matching, and other advanced methodologies are proposed in the Ontology Alignment Evaluation Initiative (OAEI) 2022 \cite{pour2023results}. Each approach attempts
to automate the matching process and alleviate the need for extensive human involvement.

With the emergence of large language models (LLMs), we have seen impressive results in many NLP downstream tasks.
Recently, using LLMs is increased for human-centric tasks, and models like ChatGPT\footnote{ChatGPT refers to ChatGPT version 4.0 unless otherwise specified.} by OpenAI\footnote{\url{https://chat.openai.com/chat}} 
have attracted attention for doing different tasks 
such as logical reasoning \cite{liu2023evaluating}, question answering \cite{tan2023evaluation}, and mental health analysis \cite{yang2023evaluations}. Prompt engineering is a skill that is required to work with LLMs efficiently. A prompt can be considered as a direction to interact with LLMs to adjust and control their output \cite{white2023prompt}. Generally, for using LLMs, there are three main approaches: fine-tuning, few-shot prompting, and zero-shot prompting. For using some LLMs in downstream tasks, fine-tuning would be helpful since it would make the LLM adapt its knowledge (from the pre-training process) to the specific task. Recently, as it is reported, models like GPT-3 \cite{brown2020language} are able to generate responses to some tasks that it has not been trained on, so prompt engineering became more popular. In few-shot prompting, a few examples of the task and the format of input/output are given to the model, so it would be able to give the output based on the format while in zero-shot prompting it is only possible to evaluate the performance of the LLM based on its knowledge in one prompt. Thus, prompt patterns are important in the results provided by these LLMS.

In this paper, we conduct a comparative analysis of ChatGPT’s performance in ontology
alignment when prompted with different strategies. We compare ChatGPT’s output with the
reference alignments provided by the Ontology Alignment Evaluation Initiative (OAEI) 2022
campaign, which uses conference-related ontologies. By evaluating ChatGPT’s performance
in a zero-shot manner, we aim to shed light on the capabilities and limitations of using a
conversational large language model for ontology matching. Furthermore, we discuss the
implications of our findings and propose potential directions for future research in this exciting
area.

\section{Methodology}
\subsection*{Data}
Our evaluation focuses on conference track ontologies provided by the OAEI \cite{zamazal2017ten}, encompassing seven ontologies: cmt, conference, sigkdd, iasted, ekaw, edas, and confOf. This selection yields 21 pairs of matched ontologies. We use the original reference alignment known as ra1\footnote{https://oaei.ontologymatching.org/2023/conference/data} for our evaluation. It is mentioned by OAEI, that M3 evaluation means both properties and classes are considered for matching. Thus, we consider ra1-M3 OAEI 2022 results for comparison. 

\subsection*{Prompts and Formatting}

An essential aspect of this evaluation involves designing prompts that effectively incorporate the triples from the conference track ontologies. We explore different approaches to include ontology triples in the prompts, with two primary methods considered: converting triples into sentences and transforming them into formatted text following the pattern Predicate(Subject, Object).

After conducting experiments and considering the effectiveness of different prompt approaches, we choose to adopt the formatted text approach for our prompts, which aligns well with suggestions from OpenAI. This formatting presents triples in a structured manner, making it easier for ChatGPT to comprehend and generate appropriate responses. For instance, an original triple such as "track subclassOf conference\_part" can be represented as "Is-a (track, conference part)" using the formatted text approach. Similarly, properties are expressed in the same structured format, such as "authorOf (Person, Document)". 

The limitation of a basic version of ChatGPT (v3.5), which we will elaborate on more in the discussion section, led us to divide it into smaller parts instead of using one long prompt. This approach allowed us to maintain essential context throughout the interaction, resulting in a better understanding of the model and more accurate responses.

In our early experiments, we found that adding more complex ontology axioms made it more difficult for ChatGPT to capture the best possible matches between two ontologies. Therefore, we decided to include only axioms that can be directly expressed as triples. We formulated our prompt with a structured approach as follows:
\\
\\

\thinkB{
\textbf{ $<$Problem Definition$>$}\\
 In this task, we are given two ontologies in the form of Relation(Subject, Object),  which \\consist of classes and properties.\\
 \textbf{$<$Ontologies Triples$>$}\\
 Ontology 1:\\
 Ontology 1 Triples\\
 Ontology 2:\\
 Ontology 2 Triples\\
 \textbf{$<$Objective$>$}\\
 Our objective is to provide ontology  mapping for the
 provided ontologies based on \\their semantic similarities.
}

Table \ref{tabexp} illustrates the diverse set prompt designs and formatting approaches used to assess ChatGPT's ontology alignment effectiveness.

\section{Results and Analysis}
In this section, we present the results of our evaluation. The objective was to gain insights and investigate this approach’s potential advantages and disadvantages. Among the prompts, "prompt 7" demonstrated the highest recall. However, it should be noted that the number of generated statements for this prompt was relatively higher than "prompt 1" since it is a repetitive prompt for each class/property name, and it tries to find the best match for each of them. Thus, the increased recall came at the cost of reduced precision, while it should be noted that some of the generated statements were deemed irrelevant even by non-expert evaluators. Nonetheless, "prompt 7" exhibited the highest F1-score among all the prompts, showcasing a balance between recall and precision.

While the first three prompts are similar in essence but have different objectives, their F1-scores are almost the same. Asking for a complete and comprehensive matching gives the highest recall, but also the least precision. On average, the first prompt achieved the best balance between recall and precision. Interestingly, employing prompts that explicitly asked for matching classes or properties, such
as prompts 4 and 5, resulted in higher recall but lower precision and F1-scores. Nevertheless,
this drawback can be mitigated by domain experts who can easily filter out irrelevant generated statements. For a more comprehensive evaluation, we compare our results with OAEI 2022 results in Table \ref{tab1}. The prompts' results are shown in Table \ref{tab2}.

\begin{table}[h]
  
  \begin{minipage}{0.65\textwidth}
    \centering
    \caption{Details of the prompts in each experiment. P\# \\shows the prompt number.}
    \resizebox{\textwidth}{!}{%
    \begin{tabular}{|c|l|l|}
    \hline
    \textbf{P\#} &
      \textbf{Description} &
      \textbf{Prompt structure} \\ \hline
    1 &
      Put all the information in a single prompt. &
      \begin{tabular}[c]{@{}l@{}}\textless{}Problem Definition\textgreater\\ \textless{}Ontologies Triples\textgreater\\ \textless{}Objective\textgreater{}\end{tabular} \\ \hline
    \multirow{2}{*}{2} &
      \multirow{2}{*}{Changing the objective of the prompts.} &
      \begin{tabular}[c]{@{}l@{}}\textless{}Problem Definition\textgreater\\ \textless{}Ontologies Triples\textgreater{}\end{tabular} \\ \cline{3-3} 
     &
       &
      \begin{tabular}[c]{@{}l@{}}Provide a complete and comprehensive\\ matching of the ontologies\end{tabular} \\ \hline
    \multirow{2}{*}{3} &
      \multirow{2}{*}{Changing the objective of the prompt.} &
      \begin{tabular}[c]{@{}l@{}}\textless{}Problem Definition\textgreater\\ \textless{}Ontologies Triples\textgreater{}\end{tabular} \\ \cline{3-3} 
     &
       &
      \begin{tabular}[c]{@{}l@{}}Match these two ontologies and provide \\ the most accurate matching you can do\end{tabular} \\ \hline
    \multirow{2}{*}{4} &
      \multirow{2}{*}{\begin{tabular}[c]{@{}l@{}}Separate the class and data/object properties\\  in two consecutive prompts.\end{tabular}} &
      \begin{tabular}[c]{@{}l@{}}\textless{}Problem Definition\textgreater\\ \textless{}Class Triples\textgreater{}\end{tabular} \\ \cline{3-3} 
     &
       &
      \begin{tabular}[c]{@{}l@{}}\textless{}Data/Object Triples\textgreater\\ \textless{}Objective\textgreater{}\end{tabular} \\ \hline
    \multirow{2}{*}{5} &
      \multirow{2}{*}{\begin{tabular}[c]{@{}l@{}}Following the Exp 2 pattern but changing \\ the objective of the prompt.\end{tabular}} &
      \begin{tabular}[c]{@{}l@{}}\textless{}Problem Definition\textgreater\\ \textless{}Class Triples\textgreater{}\end{tabular} \\ \cline{3-3} 
     &
       &
      \begin{tabular}[c]{@{}l@{}}\textless{}Data/Object Triples\textgreater\\ Match these two ontologies and provide \\ the most accurate matching you can do\end{tabular} \\ \hline
    \multirow{2}{*}{6} &
      \multirow{2}{*}{\begin{tabular}[c]{@{}l@{}}Following the Exp 2 pattern but changing \\ the order of triples to prioritizing the root \\ class entities.\end{tabular}} &
      \begin{tabular}[c]{@{}l@{}}\textless{}Problem Definition\textgreater\\ \textless{}Class Triples\textgreater{}\end{tabular} \\ \cline{3-3} 
     &
       &
      \begin{tabular}[c]{@{}l@{}}\textless{}Data/Object Triples\textgreater\\ \textless{}Objective\textgreater{}\end{tabular} \\ \hline
    \multirow{2}{*}{7} &
      \multirow{2}{*}{\begin{tabular}[c]{@{}l@{}}First, Providing the Ontologies, then asks \\ about the best class/property of the second \\ ontology that can be matched with the \\ class/property of the first one.\end{tabular}} &
      \begin{tabular}[c]{@{}l@{}}\textless{}Problem Definition\textgreater\\ \textless{}Ontologies Triples\textgreater{}\end{tabular} \\ \cline{3-3} 
     &
       &
      \begin{tabular}[c]{@{}l@{}}For a class/property in the first ontology, \\ which class/property in ontology 2 is the \\ best match?\\ \textless{}Ontology 2 Triples\textgreater{}\end{tabular} \\ \hline
    \end{tabular}
    }
    \label{tabexp}
  \end{minipage}
  \hfill
  \begin{minipage}{0.34\textwidth}
    \centering
    \vspace*{-34mm}
    \caption{\small{Comparison of OAEI 2022 \\results with ChatGPT}}
    \resizebox{\textwidth}{0.133\textheight}{%
    \begin{tabular}{|c|c|c|c|}
    \hline
    \textbf{Matcher} & \textbf{Precision} & \textbf{Recall} & \textbf{F1-score} \\ 
    \hline
    ALIN & \textbf{0.88} & 0.47& 0.61 \\ 
    \hline
    ALIOn & 0.75 & 0.22 & 0.34 \\ 
    \hline
    AMD & 0.87 & 0.43 & 0.58 \\ 
    \hline
    ATMatcher & 0.74 & 0.53 & 0.62 \\ 
    \hline
    edna & 0.79 & 0.47 & 0.59 \\ 
    \hline
    GraphMatcher & 0.8 & 0.57 & 0.67 \\ 
    \hline
    KGMatcher+ & \textbf{0.88} & 0.4 & 0.55 \\ 
    \hline
    LogMap & 0.81 & 0.58 & \textbf{0.68} \\ 
    \hline
    LogMapLt & 0.73 & 0.5 & 0.59 \\ 
    \hline
    LSMatch& \textbf{0.88} & 0.42 & 0.57 \\ 
    \hline
    Matcha & 0.38 & 0.08 & 0.13 \\ 
    \hline
    SEBMatcher & 0.84 & 0.5 & 0.63 \\ 
    \hline
    StringEquiv& 0.8 & 0.43 & 0.56 \\ 
    \hline
    TOMATO& 0.09 & 0.63 & 0.16 \\ 
    \hline
    ChatGPT-4 & 0.37 & \textbf{0.92} & 0.52 \\ 
    \hline
    
    \end{tabular}
    }
    \label{tab1}
  \end{minipage}
  
\end{table}

\section{Discussion}
Our evaluation highlighted a significant challenge related to precision. The generated statements often introduced errors that caused a decrease in precision. We identified several factors contributing to this issue:

\textbf{ChatGPT context length limit}: ChatGPT (v4.0) was used in our experiments because ChatGPT (v3.5) struggled to retain context when the input was lengthy, affecting its performance in ontology alignment tasks. ChatGPT (v4.0) has improved contextual understanding and better adaptability to long inputs, and its maximum token length of 8192 accommodates both ontology triples within the prompt.

\textbf{Inverse Functional Properties}: These Properties can lead to imprecise matches if they are not properly accounted for. For example, the statement hasBeenAssigned(Reviewer, Paper) is matched to hasReviewer(Paper, Possible\_Reviewer) by ChatGPT. However, the correct entity for this matching is ReviewerOfPaper, which is the inverse of hasReviewer. If we properly account for this inverse relationship, we can enhance precision by reducing the number of false positives.

\textbf{Matches with Subclasses}: The generated alignments sometimes matched a class in one ontology to one class and all its subclasses in the other, leading to unintended matches. For instance in the conference-edas matching, "active\_conference\_participant" and "passive\_conference\_participant" which are subclasses of conf\_participant are matched with attendee from the other ontology. Addressing this scenario is crucial for improving alignment accuracy.

\textbf{Unseen/Ambiguous Alignments}: 
Some generated alignments may appear to be accurate to non-experts, but they are actually incorrect according to reference datasets. This will be a challenge for LLMs. 
To address this issue, we propose two possible solutions: (1) revising the reference datasets to eliminate any ambiguity or inconsistency in the alignment criteria, or (2) developing a method to help LLMs detect and avoid generating implausible alignments. For instance, “camera\_ready\_paper” and “final\_manuscript” seem similar.

\textbf{Uncertain Matching}: In certain cases, even though ChatGPT acknowledges that a matching is unlikely, it still generates such matches and proposes new entities to be included in the graph.

\begin{table}
\caption{Comparison of precision, recall, and F1-score for different prompts. The cells marked with a dash (-) couldn't be completed due to token input limitations. P, R, F1 show precision, recall, and F1-score, respectively.}
\begin{adjustbox}{width=0.7\textheight,totalheight=0.3\textheight ,center}

  \begin{tabular}{|l|l|l|l|l|l|l|l|l|l|l|l|l|l|l|l|l|l|l|l|l|l|}
    \hline
    \multirow{2}{*}{Dataset} &
      \multicolumn{3}{c}{prompt 1} &
      \multicolumn{3}{c}{prompt 2} &
      \multicolumn{3}{c}{prompt 3} &
      \multicolumn{3}{c}{prompt 4} &
      \multicolumn{3}{c}{prompt 5} &
      \multicolumn{3}{c}{prompt 6} &
      \multicolumn{3}{c|}{prompt 7} \\
    & P & R & F1 & P & R & F1 & P & R & F1 & P & R & F1 & P & R & F1 & P & R & F1 & P & R & F1 \\
    \hline
     cmt-conference & 0.437 & 0.466 & 0.45 & 0.28 & 0.466 & 0.35 & 0.5 & 0.466 & 0.48 & 0.275 & 0.533 & 0.36 & 0.4 & 0.8 & 0.53 & 0.478 & 0.733 & 0.58 & 0.304 & 0.933 & 0.46\\
    \hline
    cmt-ekaw & 0.533 & 0.727 & 0.61 & 0.5 & 0.727 & 0.59 & 0.388 & 0.636 & 0.48 & 0.321 & 0.818 & 0.46 & 0.21 & 0.727 & 0.33 & 0.346 & 0.818 & 0.49 & 0.26 & 0.909 & 0.40\\
    \hline
    cmt-iasted & - & - & - & - & - & -& - & - & -& 0.173 & 1 & 0.29 & 0.266 & 1 & 0.42& 0.25 & 0.75 & 0.37 & 0.072 & 1 & 0.13\\
    \hline
    cmt-sigkdd & 1 & 0.666 & 0.8 & 0.363 & 0.666 & 0.47& 0.5 & 0.666 & 0.57 & 0.75 & 1 & 0.86& 0.625 & 0.833 & 0.71& 0.75 & 0.75 & 0.75 & 0.461 & 1 & 0.63\\
    \hline
    cmt-confOf & 0.538 & 0.437 & 0.48 & 0.36 & 0.562 & 0.44& 0.833 & 0.312 & 0.45 &0.47 & 0.562 & 0.51 & 0.4 & 0.5 & 0.44& 0.4 & 0.625 & 0.49 & 0.411 & 0.875 & 0.56 \\
    \hline
    cmt-edas & 0.666 & 0.615 & 0.64 & 0.769 & 0.769 & 0.77 & 0.562 & 0.692 & 0.62&0.529 & 0.692 & 0.6 & 0.354 & 0.846 & 0.5&0.346 & 0.692 & 0.46 & 0.28 & 0.923 & 0.43\\
    \hline
    conference-ekaw & 0.411 & 0.28 & 0.33 & 0.55 & 0.44 & 0.49 & 0.52 & 0.48 & 0.5 &0.333 & 0.52 & 0.41 & 0.344 & 0.4 & 0.37& 0.25 & 0.48 & 0.33 & 0.38 & 0.92 & 0.54\\
    \hline
    conference-iasted & - & - & - & - & - & -& - & - & - &0.285 & 0.428 & 0.34& 0.277 & 0.357 & 0.31& 0.208 & 0.357 & 0.26 & 0.325 & 0.928 & 0.48\\
    \hline
    conference-sigkdd & 0.6 & 0.4 & 0.48 & 0.379 & 0.733 & 0.5& 0.45 & 0.6 & 0.51&0.413 & 0.8 & 0.54& 0.232 & 0.666 & 0.34& 0.26 & 0.4 & 0.31 & 0.407 & 0.733 & 0.52\\
    \hline
    conference-confOf & 0.35 & 0.466 & 0.40 & 0.222 & 0.666 & 0.33& 0.4 & 0.533 & 0.46&0.357 & 0.666 & 0.46& 0.307 & 0.533 & 0.39& 0.366 & 0.733 & 0.49 & 0.466 & 0.933 & 0.62\\
    \hline
    conference-edas & 0.28 & 0.411 & 0.33 & 0.45 & 0.529 & 0.49 &0.529 & 0.529 & 0.53 & 0.375 & 0.529 & 0.44& 0.257 & 0.529 & 0.34& 0.323 & 0.647 & 0.43 & 0.35 & 0.882 & 0.50\\
    \hline
    ekaw-iasted & - & - & - & - & - & -& - & - & -&0.352 & 0.6 & 0.44& 0.222 & 0.4 & 0.28& 0.181 & 0.2 & 0.19 & 0.322 & 1 & 0.49\\
    \hline
    ekaw-sigkdd & 0.466 & 0.636 & 0.54 & 0.36 & 0.818 & 0.5& 0.411 & 0.636 & 0.5&0.28 & 0.636 & 0.39& 0.454 & 0.909 & 0.60& 0.666 & 0.727 & 0.69 & 0.33 & 1 & 0.67 \\
    \hline
    confOf-ekaw & 0.5& 0.75 & 0.6 & 0.478 & 0.55 & 0.51& 0.518 & 0.7 & 0.59& 0.355 & 0.8 & 0.49& 0.448 & 0.65 & 0.53& 0.625 & 0.75 & 0.68 & 0.558 & 0.95 & 0.70\\
    \hline
    confOf-sigkdd & 0.19 & 0.571 &0.28 &0.357 & 0.714 & 0.48& 0.235 &0.571 & 0.33&0.181 & 0.571 & 0.27& 0.23 & 0.857 & 0.36& 0.357 & 0.714 & 0.48 & 0.318 & 1 & 0.48\\
    \hline
    confOf-edas & 0.428 & 0.631 & 0.51 & 0.454 & 0.526 & 0.49& 0.428 & 0.631 & 0.51&0.363 & 0.631 & 0.46& 0.425 & 0.894 & 0.58& 0.545 & 0.631 & 0.58 & 0.444 & 0.842 & 0.58 \\
    \hline
    confOf-iasted & 0.555 & 0.555 & 0.55 & 0.461 & 0.666 & 0.54& 0.466 & 0.777 & 0.58&0.266 & 0.444 & 0.33& 0.347 & 0.888 & 0.5& 0.206 & 0.666 & 0.31 & 0.241 & 0.777 & 0.37\\
    \hline
    edas-ekaw & 0.6 & 0.391 & 0.47 & 0.423 & 0.478 & 0.45& 0.588 & 0.434 & 0.5&0.55 & 0.478 & 0.51& 0.484 & 0.695 & 0.57& 0.464 & 0.565 & 0.51 & 0.466 & 0.913 & 0.62 \\
    \hline
    edas-iasted & - & - & - & - & - & -& - & - & -&0.384 & 0.263 & 0.31&0.352 & 0.631 & 0.45& 0.307 & 0.210 & 0.25 & 0.38 & 0.842 & 0.52\\
    \hline
    edas-sigkdd & 0.5 & 0.333 & 0.4 & 0.555 & 0.666 & 0.6&0.647 & 0.733 & 0.69&0.473 & 0.6 & 0.53& 0.608 & 0.933 & 0.74& 0.444 & 0.8 & 0.57 & 0.535 & 1 & 0.7\\
    \hline
    iasted-sigkdd & 0.75 & 0.6 & 0.67 & 0.4 & 0.266 & 0.32& 0.384 & 0.333 & 0.36&0.370 & 0.666& 0.48& 0.466 & 0.466 & 0.47& 0.4 & 0.666 & 0.5 & 0.384  & 1 & 0.55\\
    \hline
    Average & 0.52 & 0.52 & 0.50 & 0.43 & 0.60 & 0.49&  0.49 &0.57  &0.51  & 0.37 &0.63 & 0.45& 0.37 & 0.69 & 0.46& 0.39 & 0.61 & 0.46 & \textbf{0.37}  & \textbf{0.92} & \textbf{0.52}\\

    \hline
  \end{tabular}
  
\end{adjustbox}
\label{tab2}
\end{table}

\section{Conclusion and Future Work}

In this paper, we have evaluated the applicability and efficiency of ChatGPT for ontology alignment using a naive approach. Our evaluation showed that ChatGPT can achieve high recall but also suffers from low precision. We identified several factors contributing to this issue, including the context length limit of ChatGPT, the handling of inverse functional properties, the matching with subclasses, unseen alignments, and uncertain matchings. 
Despite the mentioned challenges, we believe that ChatGPT has the potential to be a valuable tool for ontology alignment. The high recall of ChatGPT means that it can be used to identify a large number of potential matches, which can then be filtered by domain experts. Additionally, the ability of ChatGPT to generate new entities suggests that it could be used to expand reference ontologies.
In future work, we plan to address the precision issues identified in this paper. We also plan to explore other ways to use ChatGPT for ontology alignment, such as generating prompts for more sophisticated alignment algorithms.
Overall, we believe that the results of this paper demonstrate the potential of ChatGPT for ontology alignment. We believe that this approach can be used to improve the efficiency and effectiveness of ontology alignment tasks.

\section*{Acknowledgments}


This work was supported by the National Science Foundation (NSF) under Grant 2033521 A1. Any opinions, findings, conclusions, or recommendations expressed in this material are those of the authors and do not necessarily reflect the views of the NSF.
\\

\bibliographystyle{unsrt}  

\end{document}